\pdfoutput=1

\documentclass[11pt]{article}

\usepackage[]{acl}

\usepackage{times}
\usepackage{latexsym}
\usepackage{amsmath}
\usepackage{graphicx}
\usepackage{subfigure}
\usepackage{tabularx}
\usepackage{color}
\usepackage{multirow}

\usepackage[T1]{fontenc}

\usepackage[utf8]{inputenc}

\usepackage{microtype}

\usepackage{inconsolata}

\title{MIPS at SemEval-2024 Task 3: Multimodal Emotion-Cause Pair Extraction in Conversations with Multimodal Language Models}

\author{Zebang Cheng$^1$\thanks{Equal contributions, collaborated with CMU.}, \ \ \ \ 
Fuqiang Niu$^1$\footnotemark[1], \ \ \ \ 
Yuxiang Lin$^1$\\
{\bf \large Zhi-Qi Cheng$^2$, \ \ \ \ 
Bowen Zhang$^1$\thanks{Corresponding author.}, \ \ \ \
Xiaojiang Peng$^1$} \vspace{0.05in} \\
$^1$Shenzhen Technology University \ \ \ \ 
$^2$Carnegie Mellon University\\
\tt \small{zebang.cheng@gmail.com} \ \  {nfq729@gmail.com} 
{lin.yuxiang.contact@gmail.com} \ \ {zhiqic@cs.cmu.edu}\\
\tt \small{zhang\_bo\_wen@foxmail.com} \ \ 
{pengxiaojiang@sztu.edu.cn}\\ \\
\url{https://github.com/MIPS-COLT/MER-MCE.git}
}

\begin{document}
\maketitle

\begin{abstract}
This paper presents our winning submission to Subtask 2 of SemEval 2024 Task 3 on multimodal emotion cause analysis in conversations. We propose a novel Multimodal Emotion Recognition and Multimodal Emotion Cause Extraction (MER-MCE) framework that integrates text, audio, and visual modalities using specialized emotion encoders. Our approach sets itself apart from top-performing teams by leveraging modality-specific features for enhanced emotion understanding and causality inference. Experimental evaluation demonstrates the advantages of our multimodal approach, with our submission achieving a competitive weighted F1 score of 0.3435, ranking third with a margin of only 0.0339 behind the 1st team and 0.0025 behind the 2nd team. 
\end{abstract}

\section{Introduction}
Emotion-Cause pair extraction in conversations has garnered significant attention due to its wide-ranging applications, such as optimizing customer service interactions and tailoring content recommendations based on user emotions~\citep{xia2019emotion}. However, a fundamental challenge lies in identifying the causal determinants of emotional states~\citep{li2022emocaps}. Recent research emphasizes the exploration of causes triggering emotions from multimodal data~\citep{li2022ecpec}, followed by further generation tasks based on multimodal emotional cues~\citep{xu2024facechain}.

\begin{figure}[ht]
\centering
\includegraphics[width=0.95\linewidth]{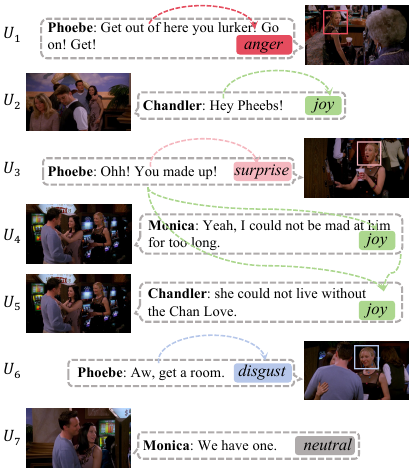}
\vspace{-2mm}
\caption{An example of an annotated conversation from the ECF dataset. Dashed lines connect each emotion label to its corresponding cause utterance, illustrating the emotion-cause pairs present in the conversation. The image modality provides additional context and cues for understanding the expressed emotions.}
\vspace{-6mm}
\label{fig:example}
\end{figure}

\begin{figure*}[ht]
\centering
\includegraphics[width=0.93\linewidth]{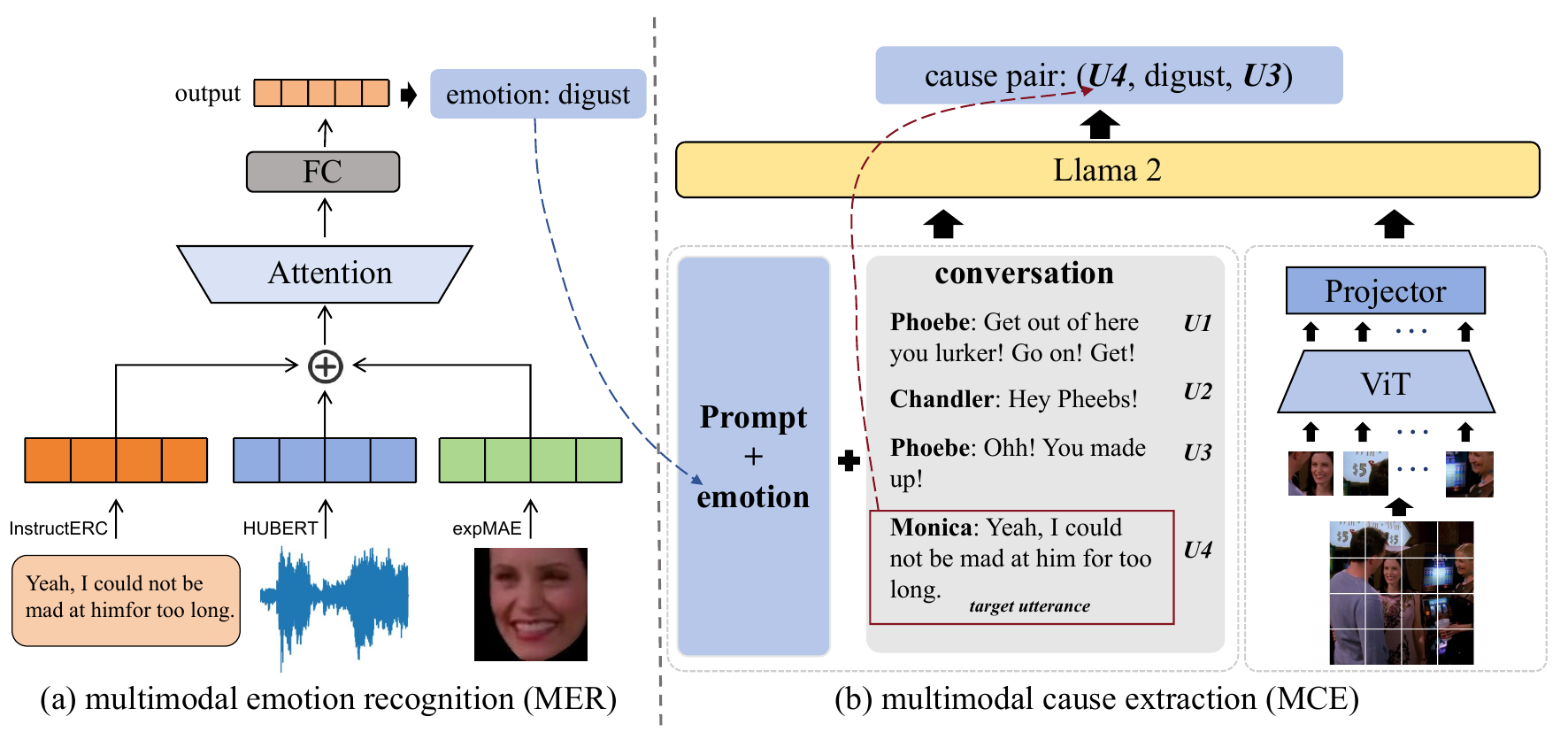}
\vspace{-2mm}
\caption{The architecture of our proposed MER-MCE framework for multimodal emotion-cause pair extraction in conversations. The framework consists of two main stages: (a) Multimodal Emotion Recognition (MER), which utilizes specialized emotion encoders to extract modality-specific features from text, audio, and visual data, and (b) Multimodal Cause Extraction (MCE), which employs a Multimodal Language Model to integrate contextual information from the conversation and visual cues to identify the utterances that trigger the recognized emotions.}
\vspace{-4mm}
\label{pipeline}
\end{figure*}

Practical conversations often exhibit multimodal cues through facial expressions, vocal changes, and textual content. Recognizing the significance of multimodal information, \citet{Emotion-Cause} proposed the task of Multimodal Emotion-Cause Pair Extraction in Conversations (MECPE) as a critical step towards understanding the fundamental elicitors of emotions. The Emotion-Cause-in-Friends (ECF) dataset, introduced in SemEval 2024 Task 3~\citep{ECAC2024SemEval}, incorporates additional modalities such as images and audio alongside the original textual data, enabling a more realistic and comprehensive approach to emotion understanding. Figure~\ref{fig:example} illustrates an example of an annotated conversation in the ECF dataset, where variations in facial expressions directly mirror the emotions expressed by the characters.

To address the MECPE task, we propose the Multimodal Emotion Recognition-Multimodal Cause Extraction (MER-MCE) framework, building upon the two-step approach introduced by \citet{Emotion-Cause}. Our method adopts a two-stage process to predict emotions and identify emotion causes in multimodal conversations. In the first stage, MER-MCE leverages text, audio, and image modalities for emotion prediction, utilizing state-of-the-art models specifically designed for capturing emotional cues from each modality. This approach sets our work apart from the first-place team in the competition, who relied solely on the textual modality for emotion recognition, and the second-place team, who employed a general-purpose model, ImageBind~\citep{girdhar2023imagebind}, for extracting visual and audio features.

In the second stage, considering the complexity of analyzing emotion causes for each utterance, we employ a Multimodal Large Language Model (LLM) to dissect the visual and textual modalities and discern the origins of each emotion. By leveraging the power of Multimodal LLMs, our approach can effectively capture the intricate relationships and dependencies present in real-world conversations, enabling a more nuanced and accurate identification of emotion causes. 

The MER-MCE framework achieved notable results in Subtask 2 of SemEval 2024 Task 3, ranking third with a weighted F1 score of 0.3435, only 0.0339 behind the first-place team and 0.0025 behind the second-place team. We evaluate the two stages of our model separately to analyze their efficacy and limitations, providing valuable insights into the inherent challenges of the MECPE task. The main contributions of this paper are as follows:
\begin{itemize}
   \vspace{-1mm}
   \item We propose the MER-MCE framework, a novel two-stage approach for Multimodal Emotion-Cause Pair Extraction in Conversations, leveraging state-of-the-art models for emotion recognition and Multimodal LLMs for cause extraction.
   \vspace{-1mm}
   \item We demonstrate the effectiveness of incorporating multiple modalities, including text, audio, and visual information, in both emotion recognition and cause extraction stages, leading to improved performance compared to approaches relying on a single modality or general-purpose feature extractors.
   \vspace{-1mm}
   \item Through comprehensive evaluation and analysis of the MER-MCE framework on the ECF dataset, we provide valuable insights into the challenges and opportunities in the field of multimodal emotion-cause pair extraction, paving the way for future research and advancements.
   \vspace{-1mm}
\end{itemize}

\section{System Overview}
\label{sec:overview}
In this work, we propose a Multimodal Emotion Recognition and Multimodal Cause Extraction (MER-MCE) framework for the task of multimodal emotion cause prediction and extraction (MECPE) in conversational settings. As illustrated in Figure~\ref{pipeline}, our MER-MCE model comprises two key modules: a multimodal emotion recognition (MER) module and a multimodal cause extraction (MCE) module, designed to work in tandem to tackle the intricate challenge of identifying emotions and their underlying causes from multimodal conversational data. Following we describe the structure of the entire system in detail.

\subsection{Multimodal Emotion Recognition}
\label{subsec:mer}
\noindent \textbf{Textual Modality}.~To comprehensively capture the rich semantic and contextual information present in real-world conversational content, our multimodal emotion recognition (MER) module adopts a carefully designed approach that leverages state-of-the-art models tailored for different modalities. For the textual modality, we employ the InstructionERC model proposed by~\citet{lei2023instructerc}, which incorporates a domain demonstration recall module based on semantic similarity to enhance feature extraction. The textual features are extracted in the form of logits, capturing the nuanced semantic representations of the conversational utterances.

\noindent \textbf{Auditory Modality}.~Recognizing the importance of auditory cues in conveying emotions, we utilize the HuBERT model proposed by~\citet{hsu2021hubert} to process the audio modality and extract hidden states as acoustic features. These acoustic features encapsulate the rich tonal and prosodic information present in the audio signals, complementing the textual and visual modalities.

\noindent \textbf{Visual Modality}.~For the visual modality, we first employ the OpenFace~\cite{baltrusaitis2018openface} open-source tool to extract facial regions from video clips, allowing our visual model to focus specifically on facial expression recognition. Subsequently, we leverage the expMAE~\cite{cheng2023semi} model to extract both static and dynamic features of facial expressions simultaneously. This dual-feature extraction approach captures the nuanced and time-varying aspects of facial expressions, which are known to be crucial indicators of emotional states.

\noindent \textbf{Multimodal Fusion Mechanism}.~To effectively fuse the complementary information from these diverse modalities, we employ an attention-based multimodal fusion mechanism, as depicted in Figure~\ref{pipeline}(a). The input features from each modality are first mapped to a common 128-dimensional space, and ReLU activation and dropout regularization are applied to introduce non-linearity and improve generalization. We then compute attention weights via dot products between the features, allowing the model to dynamically attend to the most relevant and informative cues from each modality. The resulting fused representation captures the synergistic and complementary information across modalities, enabling a comprehensive understanding of the expressed emotions.

\subsection{Multimodal Cause Extraction}
\label{subsec:mce}
Building upon the predicted emotions from the multimodal emotion recognition module, we propose a generative approach for multimodal cause extraction (MCE), harnessing the power of Multimodal Language Models (LLMs). As depicted in Figure~\ref{pipeline}(b), we adopt the MiniGPTv2~\cite{chen2023minigpt} model, a state-of-the-art multimodal LLM based on the LLaMA-2~\cite{touvron2023llama} architecture, as the backbone of our MCE module. This model is designed to integrate both visual and textual information, enabling it to extract emotional causes from multimodal conversational data.

\noindent \textbf{Image Processing}.~The image processing component of our MCE module employs the Vision Transformer (ViT)~\cite{dosovitskiy2020image} model, which divides the input image into patches and extracts visual tokens representing these patches. Notably, in contrast to the image preprocessing approach used in the multimodal emotion recognition stage, we feed the complete image to the ViT encoder. This design choice allows our model to capture comprehensive scene information and relationships among individuals, providing a holistic visual context for emotion cause extraction. The extracted visual tokens are then mapped to the textual space using a linear projection layer, facilitating the seamless integration of visual and textual information within the multimodal LLM architecture.

\begin{figure}[ht]
\centering
\includegraphics[width=0.95\linewidth]{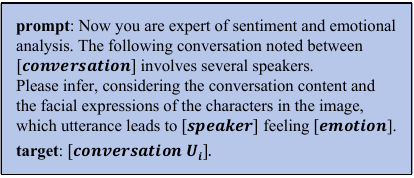}
\vspace{-2mm}
\caption{Prompt template for guiding the Multimodal LLM in sentiment analysis and emotion cause extraction from conversational data.}
\vspace{-4mm}
\label{fig:prompt}
\end{figure}

\noindent \textbf{Text Processing}.~To effectively incorporate contextual information from the conversation, we adopt a prompt-based approach in the textual processing component. The prompt template, illustrated in Figure~\ref{fig:prompt}, consists of two key elements: the prompt and the target utterance being queried. The prompt encompasses the conversation content preceding the target utterance, the speaker associated with the queried utterance, and the predicted emotion label obtained from the multimodal emotion recognition module.

\noindent \textbf{Multimodal Cause Extraction}.~The integration of image and textual data is facilitated by a trainable LLaMA2-chat (7B) model, which processes the multimodal inputs and generates natural language responses to the posed inquiries. These responses are then subjected to a similarity matching process against utterances from the historical conversation dataset. This step allows us to identify the most relevant utterance that potentially triggered the recognized emotion, culminating in the extraction of the ultimate emotion cause utterance. By seamlessly integrating visual and textual contextual information, our model can capture the intricate relationships and dependencies present in real-world conversations, enabling a more nuanced and flexible representation of emotion causes. This generative approach paves the way for more accurate and interpretable emotion cause extraction, a critical component in the overall task of multimodal emotion cause prediction and extraction.

\section{Experiments}
\label{sec:experiments}

\subsection{Experimental Setup}
We evaluate our MER-MCE model on the Emotion-Cause-in-Friends (ECF) dataset~\cite{poria2018meld}, an extension of the multimodal MELD dataset that includes emotion cause annotations in addition to the original emotion labels. The dataset provides annotations for utterances that trigger the occurrence of emotions, enabling the study of emotion-cause pairs in conversations. For Subtask 2, the labeled data is divided into "train," "dev," and "test" subsets, containing 1001, 112, and 261 conversations, respectively.

Our experimental setup involves extracting features from textual, audio, and visual modalities using various state-of-the-art pretrained models. For the textual modality, we use models such as XLNet~\cite{yang2019xlnet}, RoBERTa~\cite{liu2019roberta}, BERT~\cite{devlin2018bert}, and InstructERC~\cite{lei2023instructerc} to capture semantic and contextual information. For the audio modality, we employ models like VGGish~\cite{hershey2017vggish}, wav2vec~\cite{schneider2019wav2vec}, and HUBERT~\cite{hsu2021hubert} to extract features that encapsulate tonal and prosodic information. In the case of the visual modality, we use pretrained visual models, such as MANet~\cite{zhao2021manet}, ResNet~\cite{he2016resnet}, and expMAE~\cite{cheng2023semi}, to capture nuanced and time-varying aspects of facial expressions conveying emotional information.

\subsection{Evaluation Metrics}
The primary objective of Subtask 2 is to predict emotion-cause pairs for non-neutral categories based on the provided conversations. The performance of the participating models is evaluated using a weighted average of the F1 scores across six emotion categories: anger, disgust, fear, joy, sadness, and surprise. This weighted average F1 score provides a comprehensive evaluation of the model's ability to accurately predict emotion-cause pairs while considering the imbalanced nature of the dataset.

Further details on the dataset, experimental setup, and evaluation metrics can be found in the supplementary material.

\subsection{Emotion Recognition Analysis}
We conducted an extensive experimental evaluation of the Multimodal Emotion Recognition (MER) component within the MER-MCE framework. Table~\ref{tab:tb_emotion-results} presents the weighted F1 scores for various state-of-the-art models evaluated in the emotion recognition task, leveraging features extracted from textual, audio, and visual modalities. 

Our analysis reveals that the textual modality, which captures rich semantic information conveyed through conversations, plays a crucial role in emotion recognition. The inherent ability of the textual modality to encapsulate abstract semantic information facilitates the extraction of emotional features, resulting in higher scores compared to other modalities in unimodal emotion recognition.

\begin{table}
\resizebox{\linewidth}{!}{
    \begin{tabular}{cccc}
    \hline
    \textbf{T} & \textbf{A} & \textbf{V} & \textbf{w-avg. F1}\\
    \hline
    XLNet           & -         & -         & {0.4418}\\
    RoBERTa         & -         & -         & {0.5036}\\
    BERT            & -         & -         & {0.5128}\\
    \textbf{InstructERC} & -    & -         & \textbf{0.6606}\\
    -               & VGGish    & -         & {0.2657}\\
    -               & wav2vec   & -         & {0.4021}\\
    -      & \textbf{HUBERT}    & -         & \textbf{0.4403}\\
    -               & -         & MANet     & {0.3999}\\
    -               & -         & ResNet    & {0.4035}\\
    -      & -         & \textbf{expMAE}    & \textbf{0.4104}\\
    InstructERC     & VGGish    & -         & {0.6729}\\
    InstructERC     & HUBERT    & -         & {0.6749}\\
    InstructERC     & -         & ResNet    & {0.6774}\\
    InstructERC     & -         & expMAE    & {0.6781}\\
    -               & HUBERT    & ResNet    & {0.5113}\\
    -               & HUBERT    & expMAE    & {0.5099}\\
    InstructERC     & VGGish    & ResNet    & {0.6758}\\
    InstructERC     & VGGish    & expMAE    & {0.6779}\\
    InstructERC     & HUBERT    & ResNet    & {0.6792}\\
    \textbf{InstructERC}     & \textbf{HUBERT}    & \textbf{expMAE}    & \textbf{0.6807}\\
    \hline
    \end{tabular}
}
    \caption{Multimodal emotion recognition results}
    \vspace{-4mm}
    \label{tab:tb_emotion-results}
\end{table}

To exploit the complementary nature of different modalities, we performed multimodal feature fusion. The experimental results (Table~\ref{tab:tb_emotion-results}) demonstrate that increasing the number of modalities leads to a significant improvement in emotion recognition accuracy, empowering the model to effectively discern emotions within more complex samples. Notably, even features that exhibit relatively lower precision in unimodal emotion recognition contribute positively when integrated into the multimodal fusion framework, highlighting the importance of multimodal approaches in capturing fine-grained emotional nuances.

However, our analysis also uncovers challenges in the visual and audio modalities. In sitcoms containing multiple characters, the OpenFace tool struggles to accurately identify the current speaker, leading to noise in the visual features. Similarly, canned laughter from the audience contributes to noise in the audio modality. Consequently, models trained on these modalities exhibit subpar performance compared to the textual modality.

\subsection{Cause Extraction Analysis}
In the MCE stage, we conducted a comparison of the cause extraction capabilities between different models and the state-of-the-art MECPE-2steps model~\citep{Emotion-Cause}, with the test results presented in Table~\ref{tab:pair}. Initially, we employed the same attention model~\citep{lian2023MER2023} structure as in MER for cause extraction. However, this relatively simple model struggled to capture the relationships between utterances. We then explored the ALBEF model~\citep{li2021align} based on the transformer structure, which allowed the model to focus on the connections between different utterances. Nevertheless, limited training data and imbalanced data distribution led to overfitting.

\begin{figure}[ht] 
\centering \includegraphics[width=0.93\linewidth]{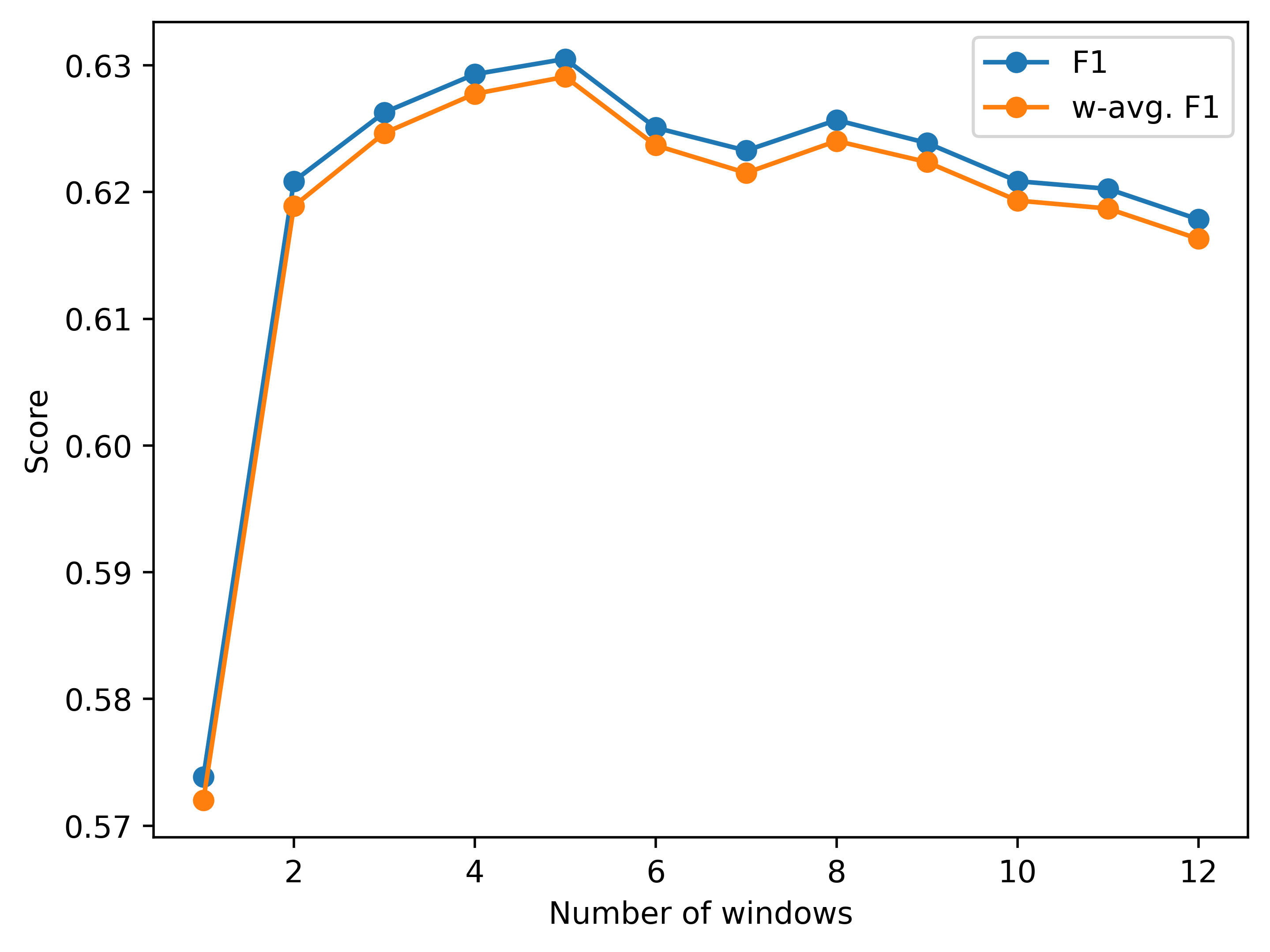} 
\vspace{-2mm}
\caption{The line graph depicting scores and historical conversation windows.} 
\vspace{-4mm}
\label{windows} 
\end{figure}

To address these challenges, we transformed the emotion cause extraction task from a traditional discriminative architecture to a generative architecture based on Multimodal LLM, resulting in improved cause extraction accuracy. We utilized the historical conversation window in the Multimodal LLM prompt to retain contextual information within the conversation. Ablation experiments were conducted to investigate the impact of varying numbers of historical conversation windows on cause extraction (Figure~\ref{windows}). To accurately assess the true influence of MCE, we employed the actual labels from the test dataset for cause extraction instead of relying solely on the emotions predicted in MER. Our experiments revealed that the effectiveness was maximized when the number of historical windows reached 5. However, as the number of windows increased further, the effectiveness gradually decreased due to the complexity of conversations with a larger historical context.

\begin{table}
\centering
\begin{tabular}{lcc}
\hline
\textbf{Method} & \textbf{F1} & \textbf{w-avg. F1}\\
\hline
Heuristic   & {-}       & {0.1864}\\
MECPE-2steps   & {-}    & {0.3315}\\
Attention   & {0.3415}  & {0.3403}\\
ALBEF       & {0.3644}  & {0.3672}\\
\textbf{MER-MCE(ours)}   & {\textbf{0.4074}}  & {\textbf{0.4042}}\\
\hline
\end{tabular}
\caption{Multimodal cause extraction results}
\vspace{-4mm}
\label{tab:pair}
\end{table}

\begin{table*}
\resizebox{\linewidth}{!}{
\renewcommand{\arraystretch}{1.28}
    \begin{tabular}{c|l|l|l}
    \hline
    \textbf{Visual Modality} & \textbf{Historical Conversation Content} & \textbf{Label Pair} & \textbf{Pred Pair}\\
    \hline
    \multirow{3}{*}{\parbox[c]{3cm}{\centering\includegraphics[width=3cm, height=1.75cm]{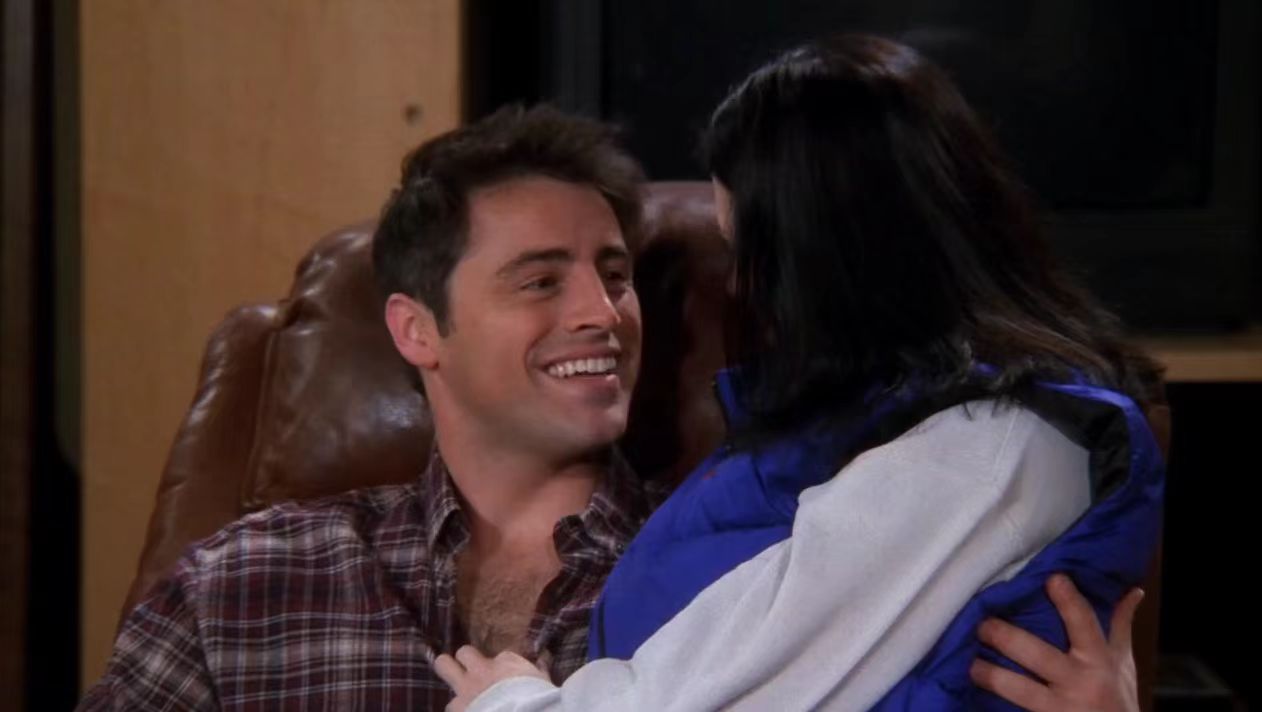}}}
               & U4: Cat.         & target: U6        & target: U6 \\
            & U5: Yes! You are so smart! I love you.         & emotion: joy          & emotion: joy\\
                & U6: I love you too.         & cause: U5         & cause: U5\\
    \hline
    \multirow{3}{*}{\parbox[c]{3cm}{\centering\includegraphics[width=3cm, height=1.75cm]{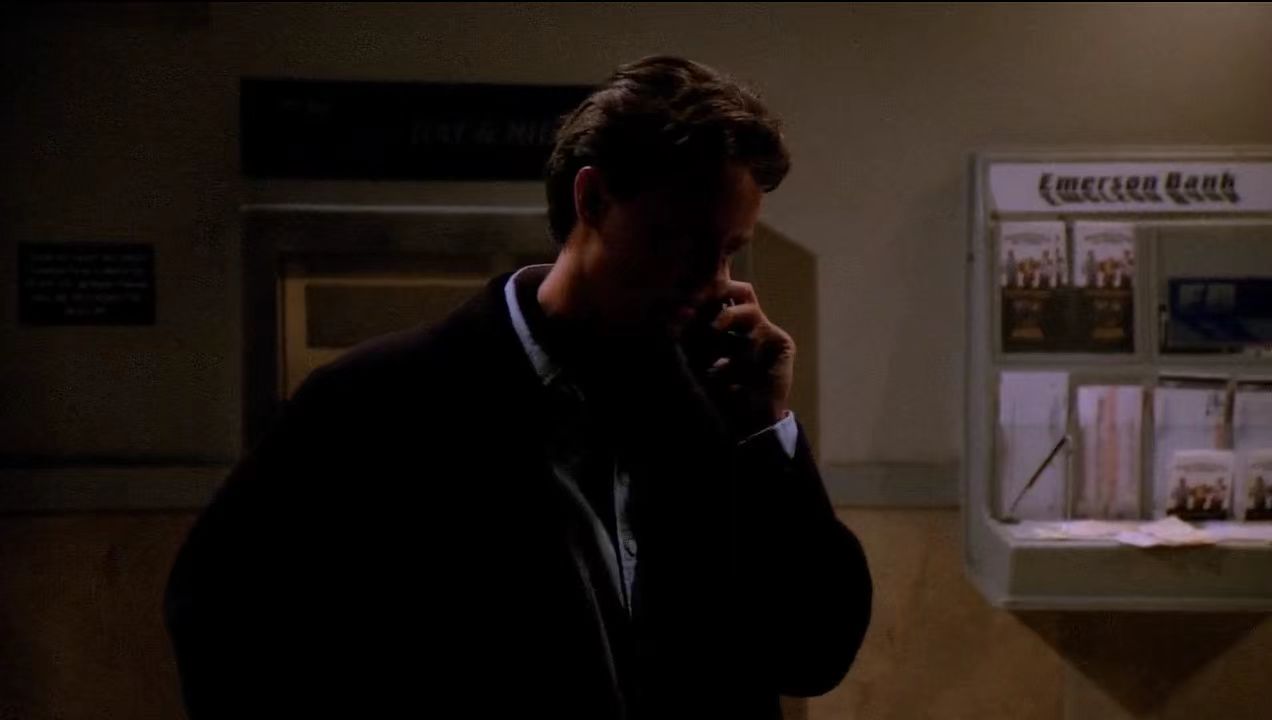}}}
               &U1: I have no idea what you just said.         & target: U2        & target: U2 \\
             & U2: Put Joey on the phone.         & emotion: anger          & emotion: anger\\
                &          & cause: U1         & cause: None\\
    \hline
    \multirow{3}{*}{\parbox[c]{3cm}{\centering\includegraphics[width=3cm, height=1.75cm]{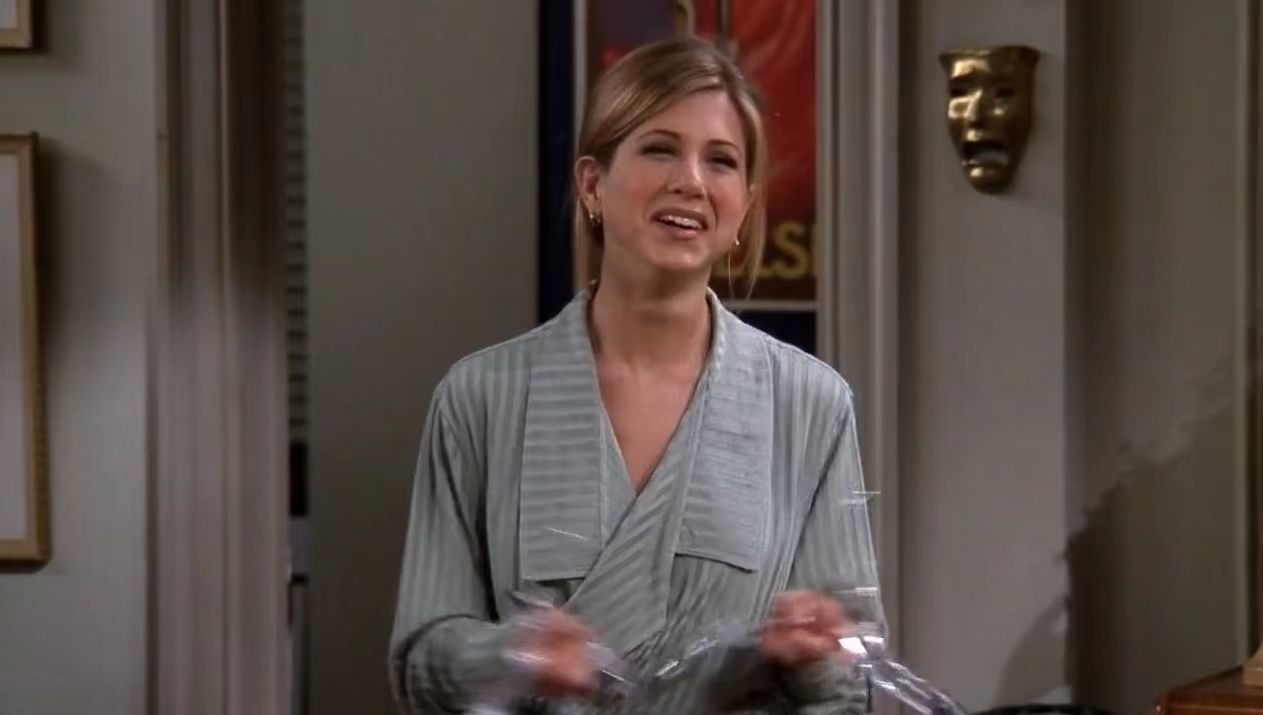}}}
           &U2: You know what? It really creeps me out ...         & target: U4        & target: U4 \\
             & U3: Sorry.         & emotion: joy          & emotion: joy\\
                & U4: I am so exited!         & cause: U5         & cause: U4\\
    \hline
    \multirow{3}{*}{\parbox[c]{3cm}{\centering\includegraphics[width=3cm, height=1.75cm]{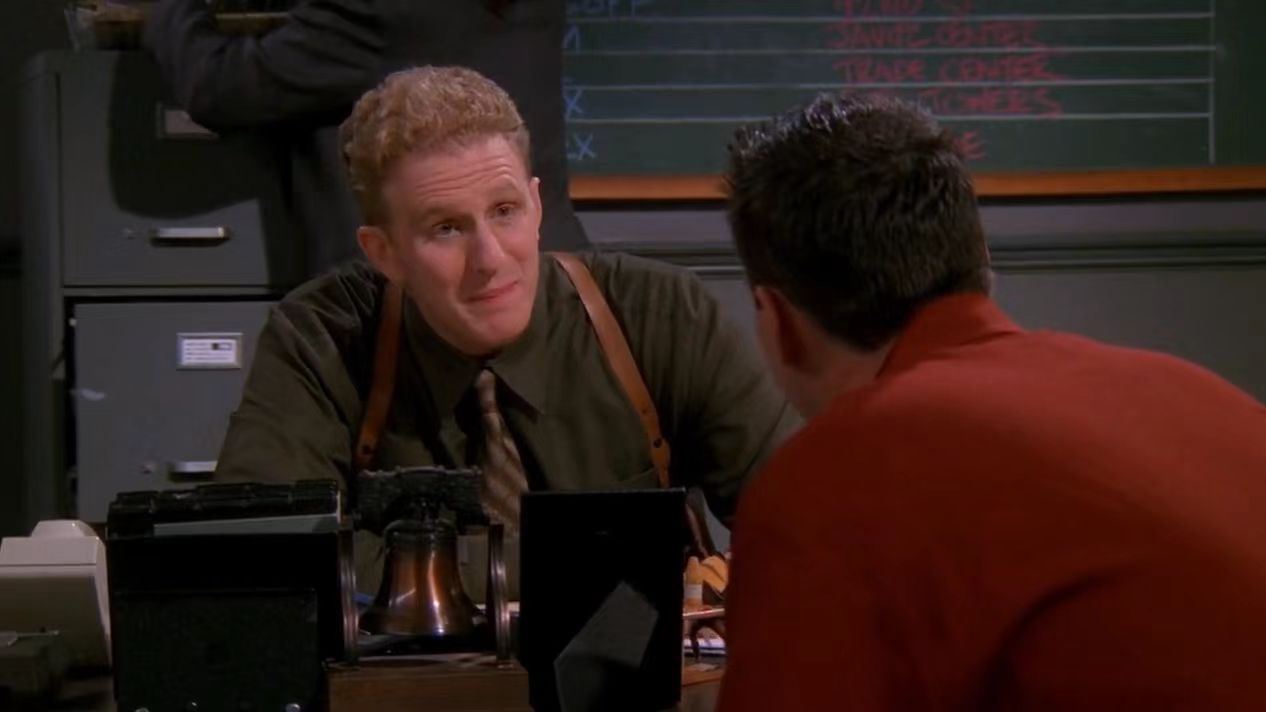}}}
           & U9:   Sure. Okay.         & target: U11        & target: U11 \\
             & U10: Uh , are you crazy? Are you insane? ...         & emotion: neutral          & emotion: joy\\
                & U11: Yeah, I ..., I just know it would make me happy.         & cause: None         & cause: U11\\
    \hline
    \end{tabular}
}
    \caption{Analysis of typical predicted emotion-cause pairs generated by the model, with emphasis on samples labeled as 'neutral' that do not have an associated emotion cause.}
    \vspace{-2mm}
    \label{tab:analysis}
\end{table*}

\subsection{Error Analysis of the Entire System}
We conducted quantitative and qualitative error analysis on the two stages of our MER-MCE framework to identify key limitations.

\begin{figure}[ht] 
\centering 
\includegraphics[width=0.5\textwidth]{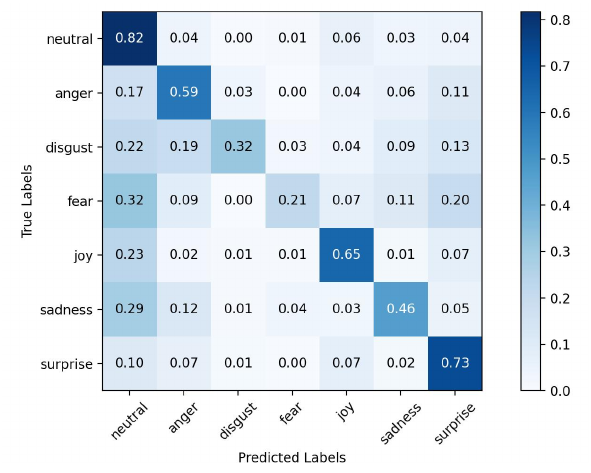} 
\vspace{-6mm}
\caption{The confusion matrix of multimodal emotion recognition result.} 
\label{confusion_matrix}
\vspace{-4mm}
\end{figure}

Analysis of the emotion recognition results using a confusion matrix (Figure~\ref{confusion_matrix}) revealed that approximately 20\% of non-neutral emotion categories were misclassified as neutral, hindering the subsequent cause analysis stage and impacting the overall recall rate. Additionally, class imbalance in the dataset adversely affected the performance of the "disgust" and "fear" categories, which had the lowest number of annotations.

Further analysis of the final Emotion-Cause Pair, based on representative samples in Table~\ref{tab:analysis}, highlighted the impact of different scenarios on our model. Facial occlusion in the visual modality (second sample) led to erroneous emotion classification, suggesting the need for more robust visual processing techniques. Strong emotional distractors in the textual modality (fourth sample) misled the model, emphasizing the importance of sophisticated language understanding methods to disambiguate distractors effectively. The real-time setting posed challenges in capturing long-range dependencies and identifying causes in future utterances (third sample), indicating the need for techniques that can model long-range dependencies and incorporate future context.

Despite these challenges, our MER-MCE model demonstrated the ability to accurately predict emotion-cause pairs by leveraging contextual information from both visual and textual modalities (first sample in Table~\ref{tab:analysis}). It identified key areas for improvement, including handling facial occlusion, disambiguating emotional distractors, and capturing long-range dependencies in real-time settings. 

\section{Conclusion}
This paper introduces the MER-MCE model for emotion cause analysis in conversations, developed for SemEval 2024 Task 3. Our model leverages multimodal information and Language Models (LLMs) to identify emotion causes in conversational data, considering textual, visual, and audio modalities. MER-MCE achieves a weighted F1 score of 0.3435, ranking third in the competition. The results demonstrate the effectiveness of multimodal approaches in capturing emotional dynamics. Future work will focus on enhancing generalizability and robustness by exploring additional modalities and advanced techniques. We plan to investigate the incorporation of pose estimagesture recognition and facial expression analysis to improve the model's ability to detect emotions.

\section{Acknowledgements}
This work was supported by the National Natural Science Foundation of China (grants 62306184 and 62176165), the Stable Support Projects for Shenzhen Higher Education Institutions (grant 20220718110918001), and the Natural Science Foundation of Top Talent of SZTU (grants GDRC202320 and GDRC202131). Zhi-Qi Cheng acknowledges support from the Air Force Research Laboratory (agreement FA8750-19-2-0200), the Defense Advanced Research Projects Agency (DARPA) grants funded through the GAILA program (award HR00111990063), and the AIDA program (FA8750-18-20018). We also appreciate the organizers of SemEval 2024.

\bibliography{acl_latex}

\appendix

\section{Appendix}
\label{sec:appendix}

\subsection{Experimental Data}
The ECF dataset, an extension of the multimodal MELD dataset~\cite{poria2018meld}, includes emotion cause annotations in addition to the original emotion labels. It provides annotations for utterances that trigger the occurrence of emotions, enabling the study of emotion-cause pairs in conversations.

For Subtask 2, the labeled data is divided into "train," "dev," and "test" subsets, containing 1001, 112, and 261 conversations, respectively. We followed this partition to train, validate, and test our MER-MCE model. To evaluate results, we submitted predictions for an additional 655 unlabeled conversations to the CodaLab platform\footnote{https://codalab.lisn.upsaclay.fr/competitions/16141}.

\subsection{Experimental Setup}
To evaluate the effectiveness of our MER-MCE model for the MECPE task, we employed various state-of-the-art pretrained models to extract features from the textual, audio, and visual modalities.

For the textual modality, we directly used the conversation text from the annotated files as input to models such as XLNet~\cite{yang2019xlnet}, RoBERTa~\cite{liu2019roberta}, BERT~\cite{devlin2018bert}, and InstructERC~\cite{lei2023instructerc} to extract textual features that capture semantic and contextual information.

For the audio modality, we extracted audio files from the video clips using FFMPEG\footnote{https://ffmpeg.org/} and fed them into models like VGGish~\cite{hershey2017vggish}, wav2vec~\cite{schneider2019wav2vec}, and HUBERT~\cite{hsu2021hubert} to obtain audio features that encapsulate tonal and prosodic information.

In the case of the visual modality, we used the OpenFace toolkit to extract facial features from the video clips while masking out the background. We then employed pretrained visual models, such as MANet~\cite{zhao2021manet}, ResNet~\cite{he2016resnet}, and expMAE~\cite{cheng2023semi}, which were initially trained on facial expression datasets, to extract visual features that capture nuanced and time-varying aspects of facial expressions conveying emotional information.

By leveraging these diverse pretrained models across multiple modalities, we aim to comprehensively capture the rich emotional cues present in the conversations and evaluate the effectiveness of our MER-MCE model in integrating these multimodal features for emotion-cause pair extraction.

\subsection{Evaluation Metrics}
The primary objective of Subtask 2 is to predict emotion-cause pairs for non-neutral categories based on the provided conversations. Each emotion-cause pair ($p_i$ = $eu_i$, $ec_i$, $cu_i$) consists of three essential elements: the index of the emotion utterance $eu_i$, the emotion category $ec_i$, and the index of the cause utterance $cu_i$.

The performance of the participating models is evaluated using a weighted average of the F1 scores across the six emotion categories: anger, disgust, fear, joy, sadness, and surprise. The F1 score for each emotion category $j$ is computed as follows:
$$F_1^j=\frac{2 \times precision^j \times recall^j}{precision^j+recall^j},$$
where $precision^j$ and $recall^j$ are the precision and recall scores for emotion category $j$, respectively.

The overall performance is determined by the weighted average F1 score, which takes into account the number of samples in each emotion category:
$$w\text{-}avg. F_1=\frac{\sum_{j=1}^{6} n^j \times F_1^j}{\sum_{j=1}^{6} n^j},$$
where $n^j$ denotes the number of samples of category $j$. This weighted average F1 score provides a comprehensive evaluation of the model's ability to accurately predict emotion-cause pairs across different emotion categories while considering the imbalanced nature of the dataset.

\end{document}